\documentclass{article}
\usepackage{ijcai23}

\usepackage{times}
\usepackage{soul}
\usepackage{url}
\usepackage[hidelinks]{hyperref}
\usepackage[utf8]{inputenc}
\usepackage[small]{caption}
\usepackage{graphicx}
\usepackage{amsmath}
\usepackage{amsthm}
\usepackage{amsfonts}
\usepackage{booktabs}
\usepackage{algorithm}
\usepackage{algorithmic}
\usepackage[switch]{lineno}
\usepackage{stfloats}

\usepackage{latexsym}
\usepackage{makecell}

\title{Find Rhinos without Finding Rhinos: Active Learning with Multimodal Imagery of South African Rhino Habitats}

\author{
Lucia Gordon$^1$\and
Nikhil Behari$^2$\and
Samuel Collier$^1$\and
Elizabeth Bondi-Kelly$^2$\and
Jackson A. Killian$^1$\and
Catherine Ressijac$^1$\and
Peter Boucher$^1$\and
Andrew Davies$^1$\And
Milind Tambe$^1$
\affiliations
$^1$Harvard University\\
$^2$Masssachusetts Institute of Technology
\emails
luciagordon@g.harvard.edu, nbehari@media.mit.edu, scollier1@g.harvard.edu, ecbk@umich.edu,
\{jkillian, catherine\_ressijac, pboucher, andrew\_davies\}@g.harvard.edu, milind\_tambe@harvard.edu}

\begin{document}

\maketitle

\begin{abstract}
Much of Earth's charismatic megafauna is endangered by human activities, particularly the rhino, which is at risk of extinction due to the poaching crisis in Africa. Monitoring rhinos' movement is crucial to their protection but has unfortunately proven difficult because rhinos are elusive. Therefore, instead of tracking rhinos, we propose the novel approach of mapping communal defecation sites, called middens, which give information about rhinos' spatial behavior valuable to anti-poaching, management, and reintroduction efforts. \textbf{This paper provides the first-ever mapping of rhino midden locations} by building classifiers to detect them using remotely sensed thermal, RGB, and LiDAR imagery in passive and active learning settings. As existing active learning methods perform poorly due to the extreme class imbalance in our dataset, we design MultimodAL, an active learning system employing a ranking technique and multimodality to achieve competitive performance with passive learning models with 94\% fewer labels. Our methods could therefore save over 76 hours in labeling time when used on a similarly-sized dataset. Unexpectedly, our midden map reveals that rhino middens are not randomly distributed throughout the landscape; rather, they are clustered. Consequently, rangers should be targeted at areas with high midden densities to strengthen anti-poaching efforts, in line with UN Target 15.7.
\end{abstract}

\section{Introduction}

\begin{figure}
\centering
\includegraphics[width=0.6\columnwidth]{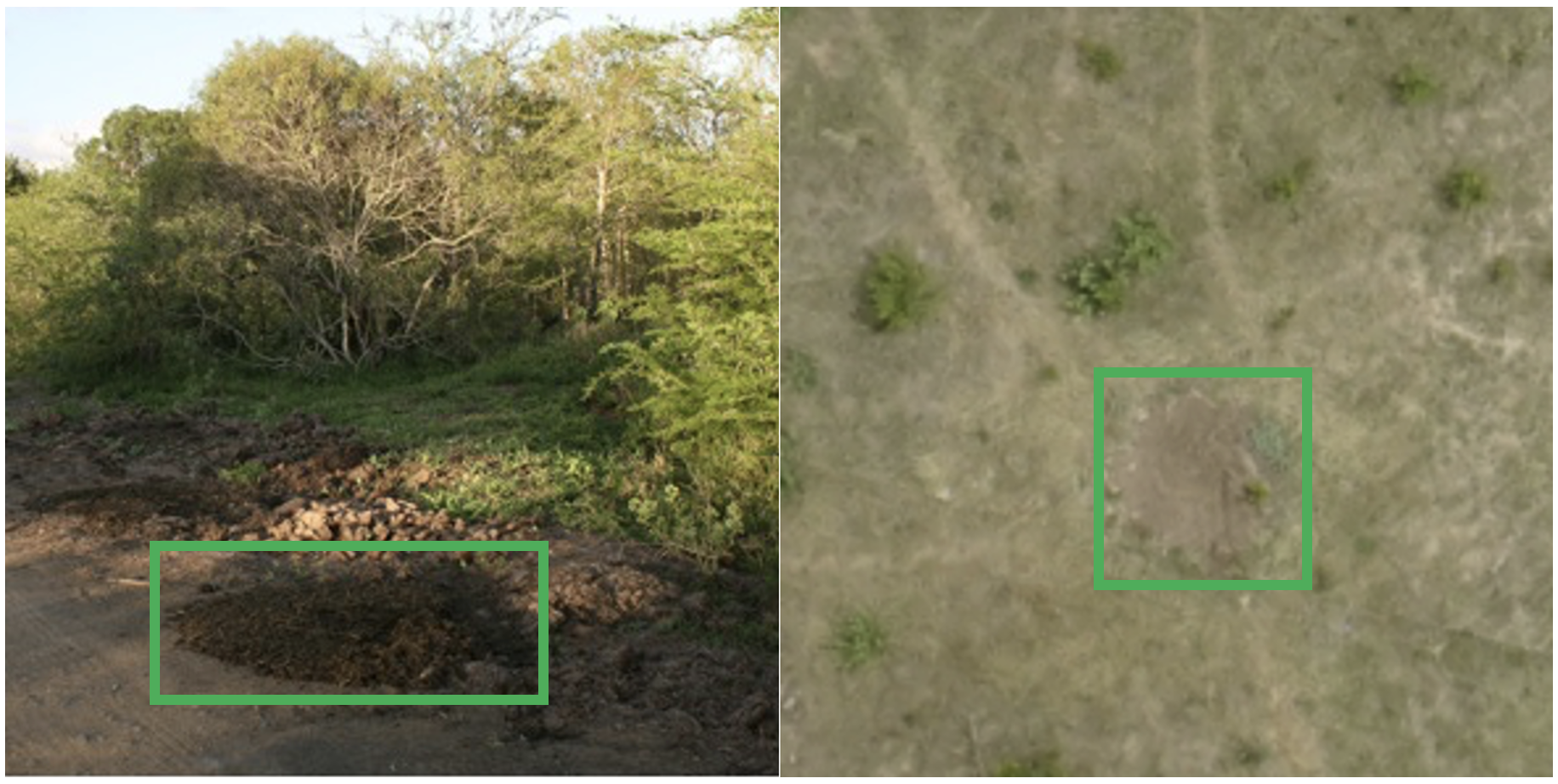}
\caption{Left: A white rhino midden next to a road in iMfolozi \protect\cite{whatisamidden}. Right: A midden in our dataset photographed by a drone. Middens boxed in green.}
\label{midden}
\end{figure}

Vertebrates are going extinct 100x faster than before the Anthropocene, indicating that the Earth is undergoing the sixth mass extinction \cite{Ceballos}. Among the human threats to biodiversity, poaching has endangered myriad species, particularly the rhino. The poaching epidemic is a barrier to achieving the United Nation's Fifteenth Sustainable Development Goal: Life on Land \cite{UN}. In particular, Target 15.7 calls for ``urgent action to end poaching and trafficking of protected species of flora and fauna" \cite{UN}. Poaching as well as habitat loss have exterminated African rhinos across much of their historical range \cite{wwf_rhinos}. In South Africa, one of the rhino's last strongholds, poaching has driven a 59\% decline in Kruger National Park's rhino population since 2013 \cite{savetherhino}. As rhinos have disproportionately large impacts on ecosystem structure and function due to their role as megaherbivores \cite{ferreira2015disruption}, it is imperative to strengthen efforts to protect them. Protecting endangered species like the rhino involves studying their patterns of habitat use \cite{johnson_gillingham_2005}, but this is challenging due to rhinos being reclusive and dangerous to observe in the wild \cite{linklater2013chemical}. Moreover, efforts to study rhinos are often constrained by the limited human and financial resources available in the African savanna landscapes rhinos inhabit, which can be large and inaccessible \cite{anderson2013lightweight}.

Rather than study rhinos directly, we propose for the first time to use rhino middens, depicted in Figure \ref{midden}, to improve our understanding of rhinos' spatial behavior. Middens are communal defecation sites used for territorial marking and social communication \cite{owen1973behavioural}. Thus, understanding midden spatial patterning is a noninvasive means for gaining insight into rhino locations and movement, providing conservation practitioners with spatial information key to effective poaching prevention, management efforts, and reintroduction plans for rhinos. Unfortunately, rhino midden locations have not yet been mapped because they are distributed throughout huge areas. Manually locating all the middens from the ground is practically impossible, and whether drones can be used for this purpose is  untested.

Against this background, this paper provides the first results from harnessing remotely sensed data to detect rhino middens along with the first spatial map of middens. Middens are considerably easier to detect in remotely sensed imagery than the elusive rhinos themselves, as the former are stationary and larger. While remote sensing technology is increasingly being used in ecological research \cite{MARVIN2016262}, the rate at which large volumes of data are collected often outpaces processing and analysis, preventing crucial insights from being gained rapidly and at scale \cite{MLconservation}.

\begin{figure}
\centering
\includegraphics[width=0.8\columnwidth]{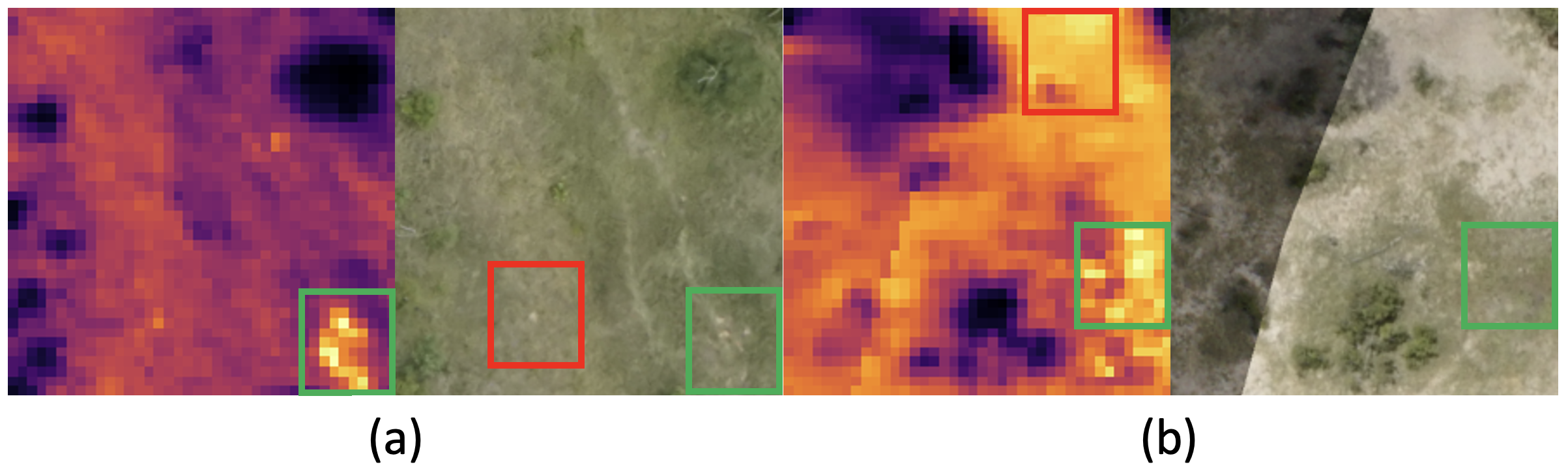}
\caption{Each pair shows the thermal (left) and RGB (right) images of the same area containing a midden. Green boxes outline middens. Red boxes outline areas that falsely appear to be middens. In (a), the midden is more obvious in the thermal image than in the RGB image, and in (b) the reverse is true.}
\label{thermalVsRGB}
\end{figure}

To overcome this challenge, we develop machine learning models utilizing thermal (heat), RGB (color), and LiDAR (light detection and ranging) imagery of a site in Kruger National Park with identified middens. We believe all three modalities can play an important role in rhino midden detection. Due to their warm temperature, middens often show up as bright areas in thermal imagery, as seen in the left image of Figure \ref{thermalVsRGB}(a). In RGB imagery, middens often appear brown, as seen in the right image of Figure \ref{thermalVsRGB}(b). LiDAR imagery is expected to be more helpful in other sites with high numbers of termite mounds because although both middens and mounds tend to be warm, the latter often have a higher slope.

First, we consider whether \emph{passive} (i.e., supervised) deep learning techniques are able to detect rhino middens in multimodal imagery. Second, we determine which data modalities and combinations thereof are most informative for automatic midden detection efforts, which is salient because of the limited resources available for conservation and ecosystem monitoring. \textbf{However, due to geographic differences between ecological sites, a deep learning model that performs well on one site may not perform well on another \cite{Beery2018}, creating a cumbersome labeling burden.} Our third contribution is therefore to develop active learning methods that strategically select images to be labeled by an expert in order to find rhino middens in an unlabeled dataset in which most images are empty. The goal is to reach an accuracy that competes with passive learning methods despite having far fewer labeled data points. However, when the dataset has extreme class imbalance, predominant active learning methods \cite{AL_pool_based,Kellenberger} are unlikely to query rare positive samples, impeding the model's learning. To overcome this challenge, we introduce the MultimodAL active learning system, which leverages information about the signal of interest to rank the instances. We then prioritize querying those most likely to be rare positive samples, accelerating the model's learning. Within MultimodAL, we also introduce an ensemble active learning strategy that dynamically weights the predictions from several models to query instances more likely to be positive samples. \textbf{Our methods apply to the general problem of identifying a rare signal of interest about which we have knowledge in an imbalanced dataset for which complete annotation is impractical.}

We train and evaluate our methods using 9,772 images of a site in Kruger National Park captured in three modalities in collaboration with South African National Parks. We perform image classification to identify midden and non-midden images and map middens geographically for the first time. We train a passive neural network on each of our data modalities (thermal, RGB, and LiDAR) as well as on fused combinations of these data types. For the middens in this site, thermal imagery is the most informative, RGB provides a slight boost in accuracy when fused with thermal, and fusing thermal with LiDAR improves recall. Next, we design and implement a novel multimodal active learning system, MultimodAL, that exploits the fact that middens are warm. We compare the performance of our query strategies against several standard baselines. MultimodAL achieves statistically identical performance to the best passive learning model with 94\% fewer labels, greatly easing the labeling burden on domain experts. Finally, mapping the rhino middens in this site reveals that they are not distributed randomly across the region but rather form clusters, so ranger patrols ought to be targeted at the areas with high midden densities. Thus, we have provided actionable information for rhino conservation as a result of our endeavor to map rhino middens rather than rhinos directly, and our methods facilitate scaling these insights to additional rhino habitats.

\section{Related Work}
We discuss related multimodal deep learning and active learning methods in this section. Several studies have fused thermal and RGB data to improve the performance of deep learning models. \citeauthor{Alexander} \shortcite{Alexander} and \citeauthor{Speth} \shortcite{Speth} utilized thermal and RGB fusion in a deep learning method to detect cracks in civil infrastructure and locate civilians in disaster zones, respectively. In our setting, we consider fusions of thermal, RGB, and LiDAR imagery. While the above works consider only passive learning settings, we also investigate multimodality in active learning environments. For our active learning models, we both evaluate performance when fusing several types of imagery beforehand and when allowing distinct models trained on different image modalities to form a ``committee" for the active learning system.

Active learning algorithms are generally distinguished by their strategy for evaluating how informative an unlabeled sample is \cite{settles.tr09}. One of the most common active learning techniques is uncertainty sampling \cite{AL_pool_based}, wherein the model requests labels for the images about which it is most uncertain. This method does not explicitly prioritize a particular class, so it is not designed to find extremely rare positive samples in a highly imbalanced dataset. In order to address this mismatch, \citeauthor{Kellenberger} \shortcite{Kellenberger} introduced positive certainty sampling, instead prioritizing images for labeling that are likely to be positive samples. Both uncertainty and positive certainty sampling were designed for a single data modality. \citeauthor{Zhang} \shortcite{Zhang} developed an active learning algorithm for thermal and RGB data that prioritizes images that are classified differently by the separate thermal and RGB models. While this method accommodates two data modalities, like uncertainty sampling it is not designed for imbalanced datasets. We introduce a form of multimodal positive certainty sampling, which prioritizes images that an ensemble of models (one for each modality) determines are likely rare positive samples. 

Because of the difficulty in identifying these rare samples, we need to make the query method as powerful as possible. An example of a method that uses expert knowledge to identify images likely to contain an object of interest is described in \citeauthor{Oliveira2} \shortcite{Oliveira2}, which uses the known human body temperature to identify people in thermal images. This technique is not, however, used within an active learning algorithm. While \citeauthor{Oliveira2} \shortcite{Oliveira2} knew the human body temperature with reasonably high precision, we face the challenge of not knowing the temperature of rhino middens. We do know, however, that they are often warmer than the surrounding ground and vegetation. Thus, in our active learning system, we prioritize images for labeling that have higher pixel values in the thermal band. With this method, the system is able to more quickly learn to distinguish between middens and non-middens.

\section{Setup}
\subsection{Data}
\begin{figure}
\centering
\includegraphics[width=0.5\columnwidth]{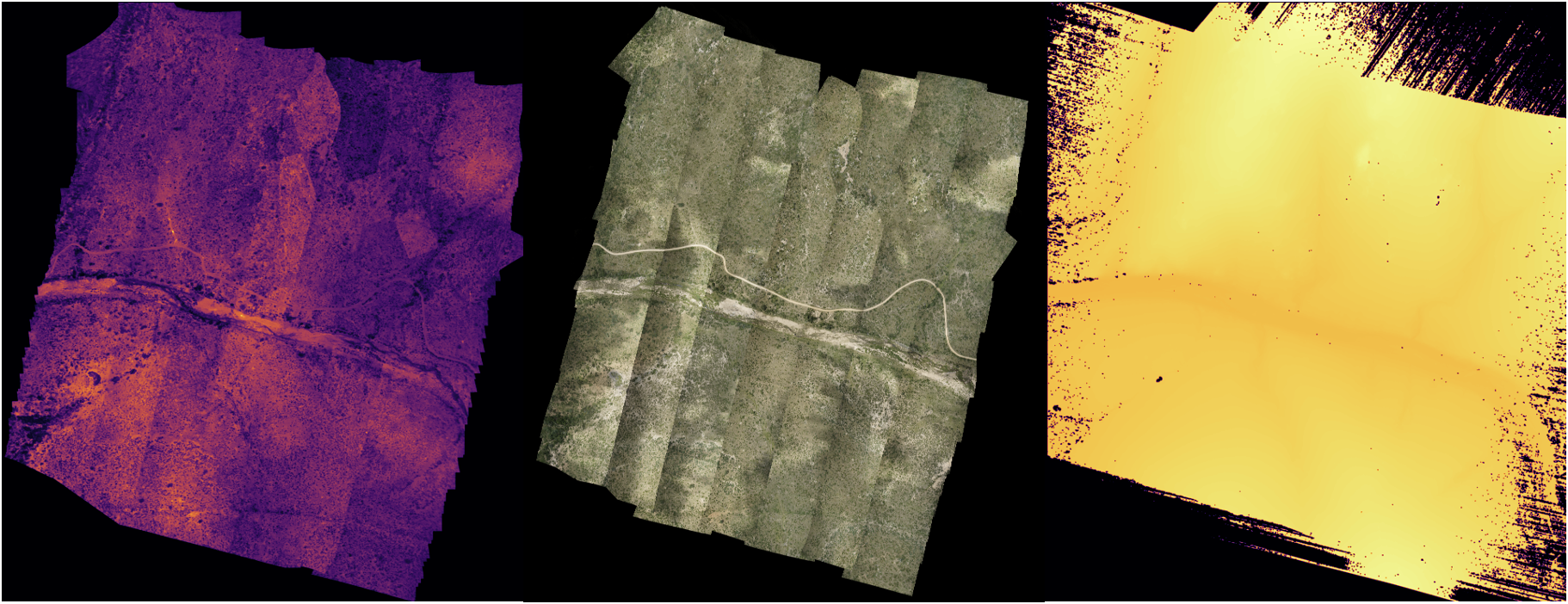}
\caption{Thermal (left), RGB (middle), and LiDAR (right) orthomosaics comprising the dataset under study.}
\label{orthomosaics}
\end{figure}

The remote sensing data used for this project was collected by a DJI M600 multicopter flown over a 284-hectare site in Kruger National Park in January 2020. This drone was equipped with an animal landscape observatory sensor package, consisting of a FLIR Tau-2 thermal camera, a Sony A6000 camera, and a Riegl VUX-1LR LiDAR scanner, which simultaneously collected thermal, RGB, and LiDAR data, respectively, throughout the drone's flight. The thermal imagery was rectified and mosaicked at a resolution of 0.5 m, the RGB imagery at 0.05 m, and the LiDAR imagery at 0.25 m, yielding the orthomosaics shown in Figure \ref{orthomosaics}.

The ecologists on our team identified candidate middens in the thermal and RGB orthomosaics and confirmed their presence on the ground, yielding a list of the $x$ and $y$ coordinates of the centers of 52 rhino middens. We mapped these middens onto the orthomosaics and then cropped them using an interval of 20 m (40 pixels for thermal, 400 for RGB, and 80 for LiDAR) and a stride of 5 m (10 pixels for thermal, 100 for RGB, and 20 for LiDAR). Each cropped image was assigned a label of 1 if it contained the center of a midden and 0 otherwise. We downshifted the pixel values of each thermal image such that the cropped thermal images all had a minimum of 0 to enable a meaningful comparison among them. After removing images with all zeros in either the thermal or RGB bands, we were left with 89 images with middens and 9,683 empty images, which means that our dataset has 9,772 images (in each modality), 0.91\% of which have a midden. We fused the images using the blend function in the PIL class, yielding fusions of thermal and RGB, thermal and LiDAR, RGB and LiDAR, and thermal, RGB, and LiDAR images, with each data modality weighted equally.

\subsection{Model}
We employ transfer learning with a VGG16 model pretrained on the ImageNet dataset \cite{vgg16}. We freeze all the parameters in the model except for those in the classifier. We alter the final linear layer to have a single out feature and then end with a sigmoid function so that the output of the model represents the probability that an image contains a midden. For all our models, we use a batch size of 10, the Binary Cross Entropy loss function in PyTorch, and an Adam optimizer with a learning rate of 0.0001.

\section{Active Learning Methodology}
\subsection{MultimodAL Algorithm} \label{subsec:multimodal_algorithm}
Active learning aims to reduce the number of instances that need to be labeled to train a model by requesting labels for those that are most useful for its learning. The general procedure works as follows: (i) a small batch of labeled instances is used to begin training a model, (ii) the model then uses some criteria to select the next batch of instances to be labeled, typically those about which the model is least certain \cite{AL_pool_based}, (iii) this process continues until a labeling budget is reached, and (iv) the trained model can then be used for inference on the remaining unlabeled instances.

Problematically, however, traditional active learning approaches can have poor performance in the presence of severe class imbalance. To address this challenge, our active learning algorithm, MultimodAL, is designed to detect as many of the rare positive samples as possible in each round. To achieve this, rather than have the model predict on the entire training dataset as is done in many typical active learning systems, we propose constraining the set of instances on which the model predicts through a novel technique that exploits some characteristic of the object of interest that can be used for ranking. Furthermore, we propose a dynamic method for combining the outputs of several models trained on different data modalities to further speed up learning.

\begin{figure}
\centering
\includegraphics[width=0.75\columnwidth]{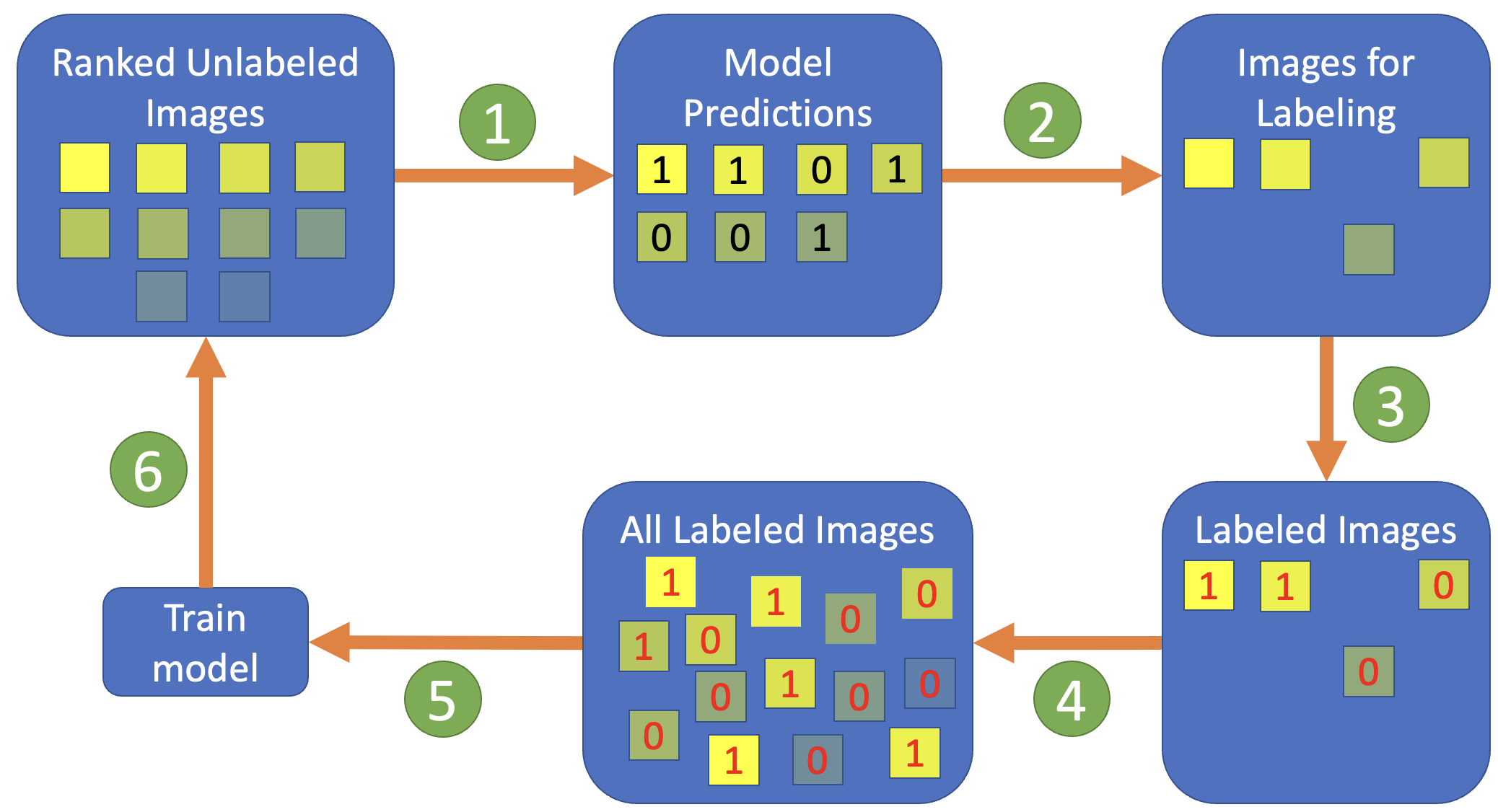}
\caption{Active learning cycle where the images are ranked by their brightness. (1) Predict on highest-ranked images. (2) Query images predicted to be positive. (3) Assign labels to queried images. (4) Add the newly labeled images to the set of all labeled images. (5) Train the model on a selection of the labeled images. (6) Restart.}
\label{diagram}
\end{figure}

We first describe our method assuming a single data modality, diagrammed in Figure \ref{diagram}. We assume each instance in the dataset can be assigned a value corresponding to a \emph{metric} (e.g. temperature, color, etc.) that is associated with the desired rare signal of interest. We then rank all the instances by the distance between their value of the informative metric and a specified \emph{target} value (e.g. human body temperature, color of grass, etc.) characteristic of the desired signal (top left box of Figure \ref{diagram}). Once the images are ranked, we select a subset to be labeled. Let $b$ be the size of a batch of images selected by the active learning system for labeling. (1) To produce each of the batches, we first compute the output of the model on the sample with the highest ranking out of those remaining unlabeled. To reflect the uncertainty captured in the model's output, we do not simply assign the instance the highest probability class. Instead, we classify it by randomly sampling a class according to the model's output (i.e., the output specifies the parameters of a multinomial distribution). If the instance is ultimately predicted to be a positive sample, we add it to the batch. We then feed the sample with the next highest ranking to the model and continue this process until we have a batch with $b$ samples predicted by the model to be positive. In this way, we bias the model towards selecting high-ranking images that we know are more likely to be positive instances. (2) Next, the batch is sent to the labeler. (3) The batch is labeled by the annotator and then (4) added to the set of instances queried so far. Because positive samples may be so rare, batches can be imbalanced toward the negative class(es). (5) If all the instances queried so far are negative, then we select all of them for training. If any are positive, we take all of those for training and randomly select an equivalent number of negative instances to get a balanced training set. At this point the model weights are reset to their initial values to prevent overfitting on a small labeled dataset, and the model is then trained on the selected instances. (6) This process is repeated until we exhaust a budget on the number of labels that can be provided.

\begin{figure}[b]
\centering
\includegraphics[width=0.5\columnwidth]{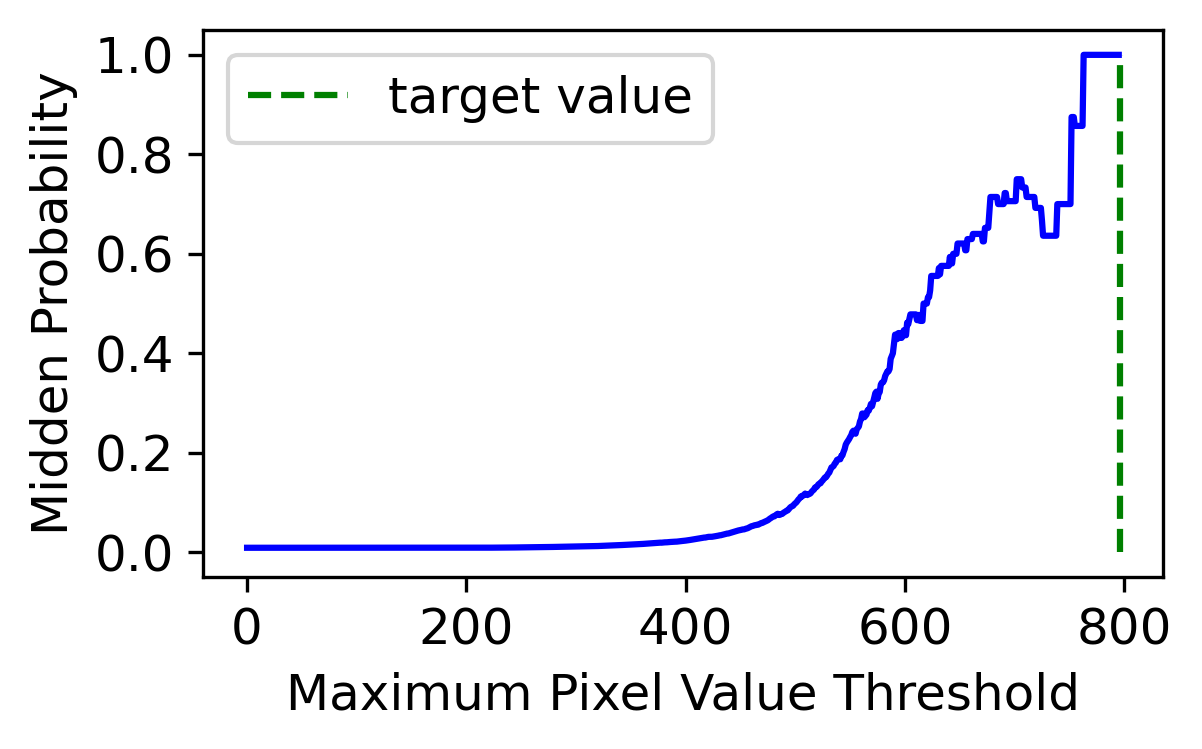}
\caption{Probability that an image contains a midden tends to increase with its maximum thermal pixel value.}
\label{middenprobplot}
\end{figure}

Since we are also interested in settings with multiple data modalities, we develop a modification of the above procedure to accommodate an ensemble of models, each of which is trained on its own data modality. Specifically, we modify the way in which the prediction for each instance is calculated. Above, the prediction was simply extracted from the output of a single model, where the output contains the probabilities that the instance belongs to each of the possible classes. In the multimodal setting, we have the outputs of multiple models with possibly differing accuracies, so we want to weight their predictions accordingly. In particular, we assign each instance a score for each of the possible classes given by a weighted sum of the models' outputs. If there are $M$ models, then the score for an instance belonging to class $i$ is calculated using Equation \ref{score}. Then as in the unimodal case, the prediction for the instance is obtained by sampling from a multinomial distribution with the class scores as inputs.

\begin{equation}
    \text{score}_i = \sum_{m=1}^M \text{weight}_m\times\text{output}_{m,i}
\label{score}
\end{equation}

After being initialized to $1/M$, the weights are updated in each subsequent round according to the number of queried instances that the models have classified correctly. The weight for model $m$ is given by Equation \ref{weights}, where correct$_m$ is the number of instances queried so far that were classified correctly by model $m$. By construction the weights sum to 1, so the resulting scores can be interpreted as probabilities.

\begin{equation}
    \text{weight}_m=\frac{\text{correct}_m}{\sum_{n=1}^M\text{correct}_n}
\label{weights}
\end{equation}

\subsection{Intuition for Ranking Idea}
Having described our active learning algorithm, we now present additional analysis that provides intuition for our choice of ranking metric for our dataset. Note that the following analysis is \emph{not} necessary to use the algorithm and is solely for explanatory purposes.

For our setting, we have chosen the maximum thermal pixel value as the metric to be used for ranking in descending order. This sets the target as the maximum pixel value across all of the thermal images, exploiting our knowledge that the sought-after rhino middens are warm. To demonstrate that this ranking technique effectively prioritizes midden images for labeling, we compute the probability of a thermal image containing a midden given that its maximum pixel value (MPV) is no less than a threshold $t$. To calculate this, we first apply Bayes' rule, shown in Equation \ref{bayesrule}.

\begin{equation}
    \mathbb{P}(\text{midden}|\text{MPV}\geq t)=\frac{\mathbb{P}(\text{MPV}\geq t|\text{midden})\mathbb{P}(\text{midden})}{\mathbb{P}(\text{MPV}\geq t)}
\label{bayesrule}
\end{equation}

Let $m$ be the total number of middens and $m_t$ be the number of midden images with MPV no less than $t$. The first factor in the numerator is then $\mathbb{P}(\text{MPV}\geq t|\text{midden})=\frac{m_t}{m}$. Plugging this into Equation \ref{bayesrule} gives Equation \ref{middenprob}.

\begin{equation}
    \mathbb{P}(\text{midden}|\text{MPV}\geq t)=\frac{m_t\times \mathbb{P}(\text{midden})}{m\times\mathbb{P}(\text{MPV}\geq t)}
\label{middenprob}
\end{equation}

We plot Equation \ref{middenprob} in Figure \ref{middenprobplot}, which shows that the probability of an image containing a midden tends to increase as its maximum pixel value nears the target value, demonstrating the utility of the ranking method. \textbf{More generally, we expect any dataset with an appropriately chosen informative metric and target to obey a similar pattern: the probability of being a positive instance drops with increasing distance from the target value.} Our ranking-based active learning query strategy is designed for any such dataset.

\section{Results}
In this section, we present the performance of models passively trained on thermal, RGB, LiDAR, and fused imagery and show the trained models are able to detect middens in a held-out test set with high accuracy. We also display the performance of our MultimodAL active learning algorithm in comparison to several baselines. We demonstrate our method efficiently selects images for labeling and achieves fast midden retrieval in a large, imbalanced, and initially unlabeled dataset. All error bars in Subsections \ref{passive} and \ref{active} show one standard error of the mean in each direction. All models are trained for 10 epochs, and a threshold of 0.5 is used on the models' sigmoid output for test image classification. Each experiment is run 30 times.

\subsection{Detecting Middens with Passive Learning}\label{passive}

\begin{figure}
\centering
\includegraphics[width=0.9\columnwidth]{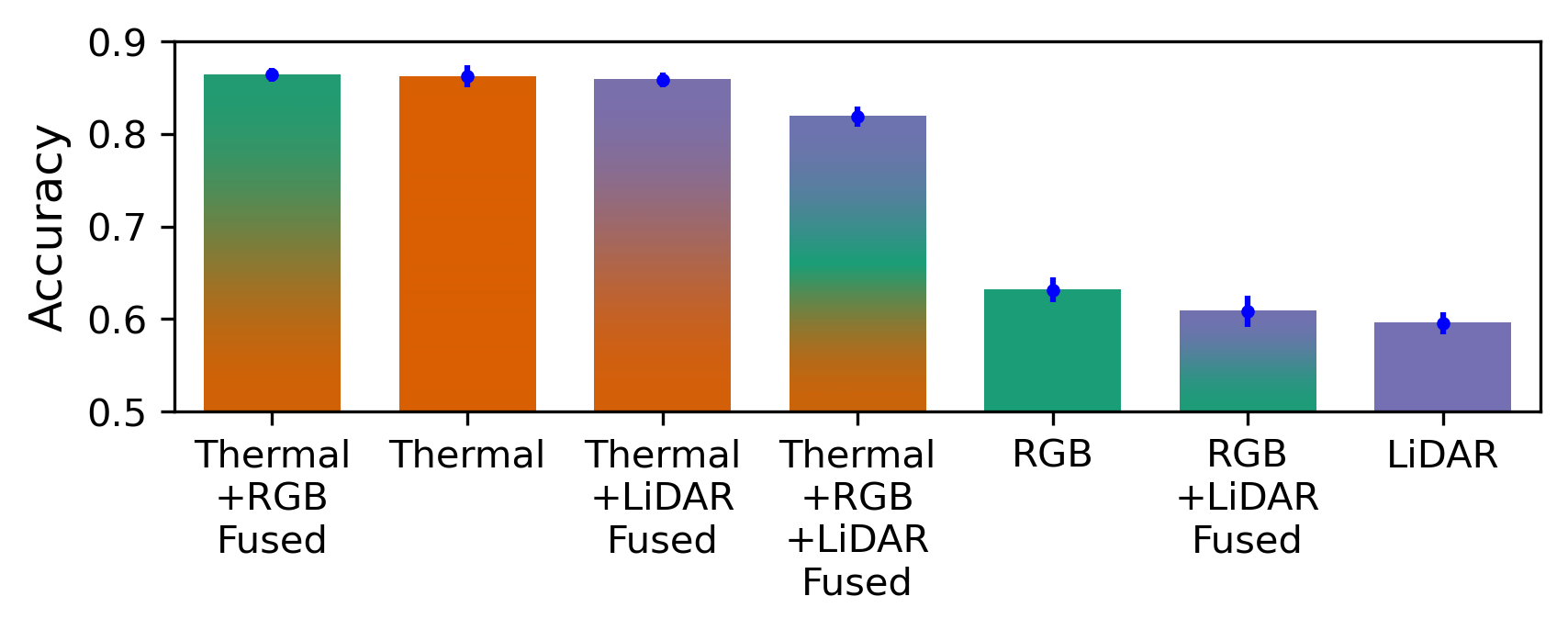}
\caption{Mean accuracy for the passive models across 30 trials after training for 10 epochs. Models are ranked in descending order by accuracy.}
\label{passivefig}
\end{figure}

\begin{table*}[bp]
\centering
\begin{tabular}{|c|c|c|c|c|c|c|c|}
\hline
    & \makecell{T+R Fused} & Thermal & \makecell{T+L Fused} & \makecell{T+R+L Fused} & RGB & \makecell{R+L Fused} & LiDAR \\ 
\hline
    Accuracy & \boldmath$.864\pm.008$ & $.862\pm.012$ & $.858\pm.008$ & $.818\pm.011$ & $.632\pm.014$ & $.608\pm.016$ & $.595\pm.012$ \\
\hline
    Precision & \boldmath$.875\pm.010$ & $.846\pm.014$ & $.826\pm.011$ & $.823\pm.012$ & $.629\pm.015$ & $.616\pm.020$ & $.632\pm.023$ \\
\hline
    Recall & $.855\pm.016$ & $.894\pm.016$ & \boldmath$.917\pm.011$ & $.818\pm.017$ & $.670\pm.023$ & $.678\pm.028$ & $.576\pm.041$ \\
\hline
    F1 & $.861\pm.009$ & $.866\pm.012$ & \boldmath$.866\pm.007$ & $.817\pm.012$ & $.642\pm.015$ & $.631\pm.015$ & $.565\pm.024$ \\
\hline
\end{tabular}
\caption{Passive learning statistics. T=Thermal, R=RGB, and L=LiDAR. The Thermal+RGB (T+R) Fused model achieves the best performance on accuracy and precision, and the Thermal+LiDAR (T+L) Fused model achieves the best recall and F1 score.}
\label{passivetable}
\end{table*}

We passively train models with images in different modalities to establish that neural networks are capable of accurately detecting rhino middens in remotely sensed imagery. To train our passive learning models we assume that the system has access to all of the images' labels from the start. We split this labeled data into training and test sets with the following random selections. First, we add 80\% of the midden images (71) to the training set and leave 20\% (18) for the test set. We then add 18 empty (non-midden) images to the test set, yielding a balanced test set of 36 images. Of the remaining empty images, we randomly sample 71 and add them to the training set to balance it, yielding a balanced training set of 142 images.

For each trial, we train the model on the thermal, RGB, LiDAR, or fused imagery and then record the accuracy on the test set at the end, graphed in Figure \ref{passivefig}, where each trial has a different random assignment of images to the training and test sets. We also report the mean and standard errors of the accuracy, precision, recall, and F1 score across the passive trials in Table \ref{passivetable}. We observe that the Thermal+RGB Fused model achieves the best performance on accuracy and precision, and the Thermal+LiDAR Fused model achieves the best recall and F1 score. Among the three data modalities, the Thermal model significantly outperforms the RGB and LiDAR models on our dataset. Thus, based on our results, if resources on the ground are very constrained, it could be helpful to prioritize thermal sensors for midden detection.

\subsection{Efficient Querying with Active Learning}\label{active}
The passive learning systems achieve high accuracy but require 9,736 labeled images, so such systems \emph{cannot} be used to map middens in other sites where we do not have any of their coordinates beforehand. Moreover, differing topography among sites makes transfer learning challenging. Labeling enough images for passive learning systems to be accurate would take domain experts dozens of hours. In this subsection we demonstrate that our MultimodAL algorithm, described in Subsection \ref{subsec:multimodal_algorithm}, is able to match the best passive learning performance with 94\% fewer labeled images by employing an efficient query strategy. While we do have the ground-truth midden locations for this site, enabling us to test various detection methods, we use the results hereinafter as a proof of concept for our system.

For active learning we use the same dataset and perform the same train-test split procedure as for passive learning except that we do not balance the training set, so all the empty images not set aside for testing (9,665) are available to the system during training. We test two versions of MultimodAL. First, MultimodAL: Thermal+RGB Fused uses the version of our proposed algorithm for a single model trained on the pre-fused thermal+RGB images, and second, MultimodAL: Thermal + RGB maintains separate models trained on the thermal and RGB images and uses the version of the algorithm that accommodates multiple modalities. In addition to our proposed MultimodAL algorithm, we implement several baselines: (1) Random: randomly chooses images to be labeled in each round, (2) Uncertainty: requests labels for the images with model outputs closest to 0.5, a very common active learning strategy \cite{AL_pool_based}, (3) Positive Certainty: requests labels for the images with outputs closest to 1, an active learning strategy more tailored to imbalanced datasets like ours  \cite{Kellenberger}, and (4) Disagree: asks the user to label the images with the greatest difference in output between the thermal and RGB models \cite{Zhang}. We train and evaluate the single-model systems on the thermal+RGB fused data since that model performed the best in the passive setting. For the multi-model systems, we use separate thermal and RGB models. We train each of these on the images in its modality. In evaluating the system as a whole, images in the test set are classified by weighting the models' outputs by their weights learned during training, calculated using Equation \ref{weights}.

\begin{figure}
\centering
\includegraphics[width=0.9\columnwidth]{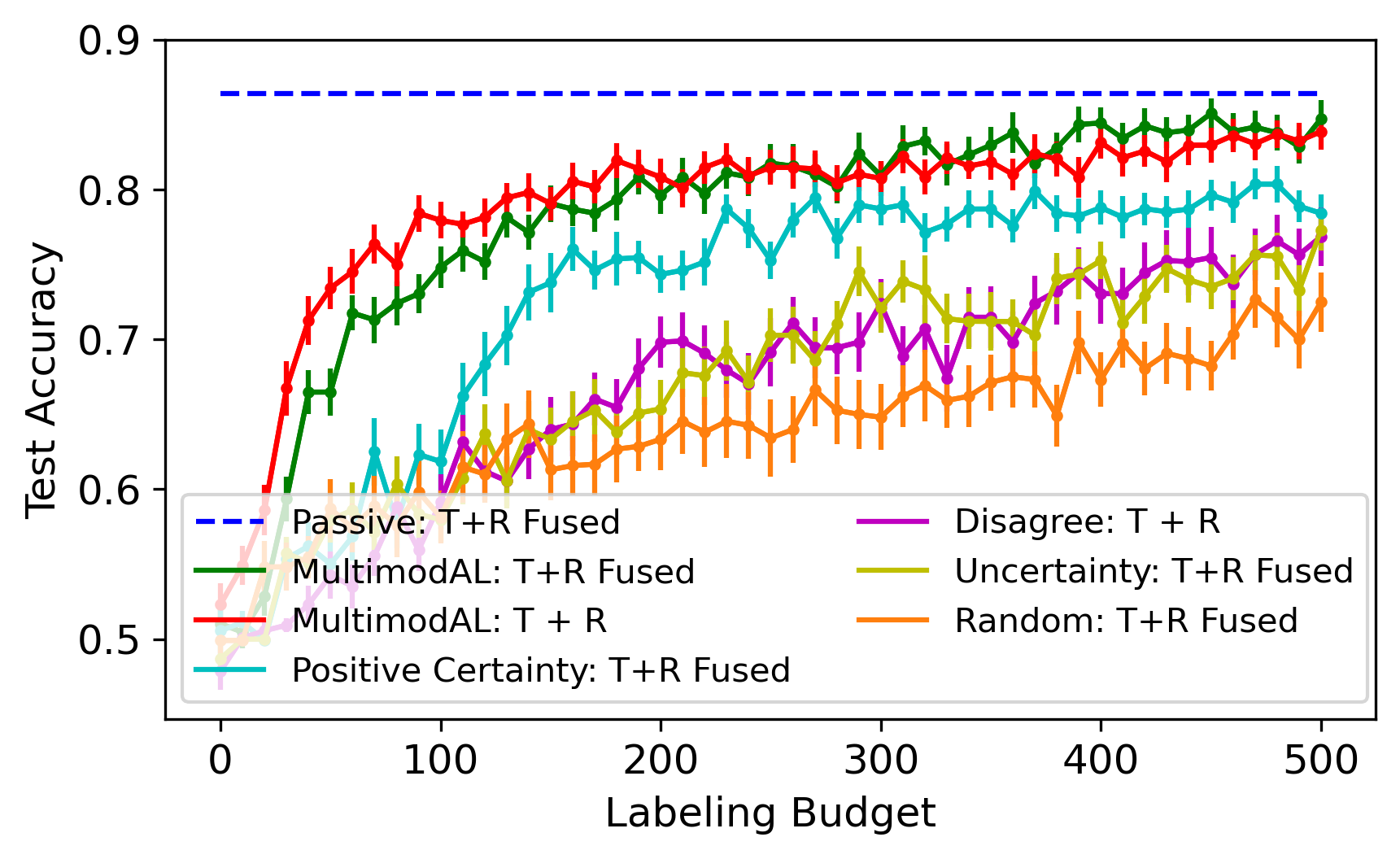}
\caption{Mean accuracy on the test set for the active learning methods with up to 500 images labeled across 30 trials. Both versions of MultimodAL outperform the baselines and with 500 labels are statistically indistinguishable from the best passive method.}
\label{activelearning}
\end{figure}

We graph the accuracy of our proposed methods in comparison to the baselines' as the labeling budget increases in Figure \ref{activelearning}. We find that both versions of MultimodAL outperform all of the baselines by a statistically significant amount ($p<0.05$) at 500 labels. This is despite the fact that Positive Certainty is statistically significantly better ($p<0.05$) than Random, demonstrating it is a strong baseline. Positive Certainty outperforming the other baselines is consistent with our hypothesis that prioritizing images the system thinks are positive samples is more suitable for highly imbalanced datasets than prioritizing images about which the system is uncertain. Even so, it is the two variants of MultimodAL that get within 0.026\% of the best passive learning accuracy by 500 labels, which is \emph{not} a statistically significant difference ($p>0.05$). Hence, there is a negligible difference in performance despite the 94\% difference in the number of labeled images provided to the passive system and MultimodAL. 

Interestingly, MultimodAL: Thermal + RGB outperforms MultimodAL: Thermal+RGB Fused for small numbers of labels ($<200$), despite the opposite pattern holding for larger numbers of labels. Because MultimodAL: Thermal + RGB combines the scores across two modalities, it will more strongly prioritize images that are very clearly middens in both imagery types than the single-model MultimodAL: Thermal+RGB Fused, which will assign a high score to images that are clear in \emph{either} data type. This could result in more informative midden images being selected in the former case and leading to improved performance when the number of discovered middens is low. Overall, the advantages of MultimodAL are greatest when the labeling budget is very small ($<150$). For example, we see that for up to 50 images labeled only MultimodAL does better than Random. Furthermore, we see diminishing returns in accuracy for MultimodAL after around 200 labels once the system has been trained on enough middens ($\sim33$) to serve as a reasonable detector.

\begin{figure}
\centering
\includegraphics[width=0.9\columnwidth]{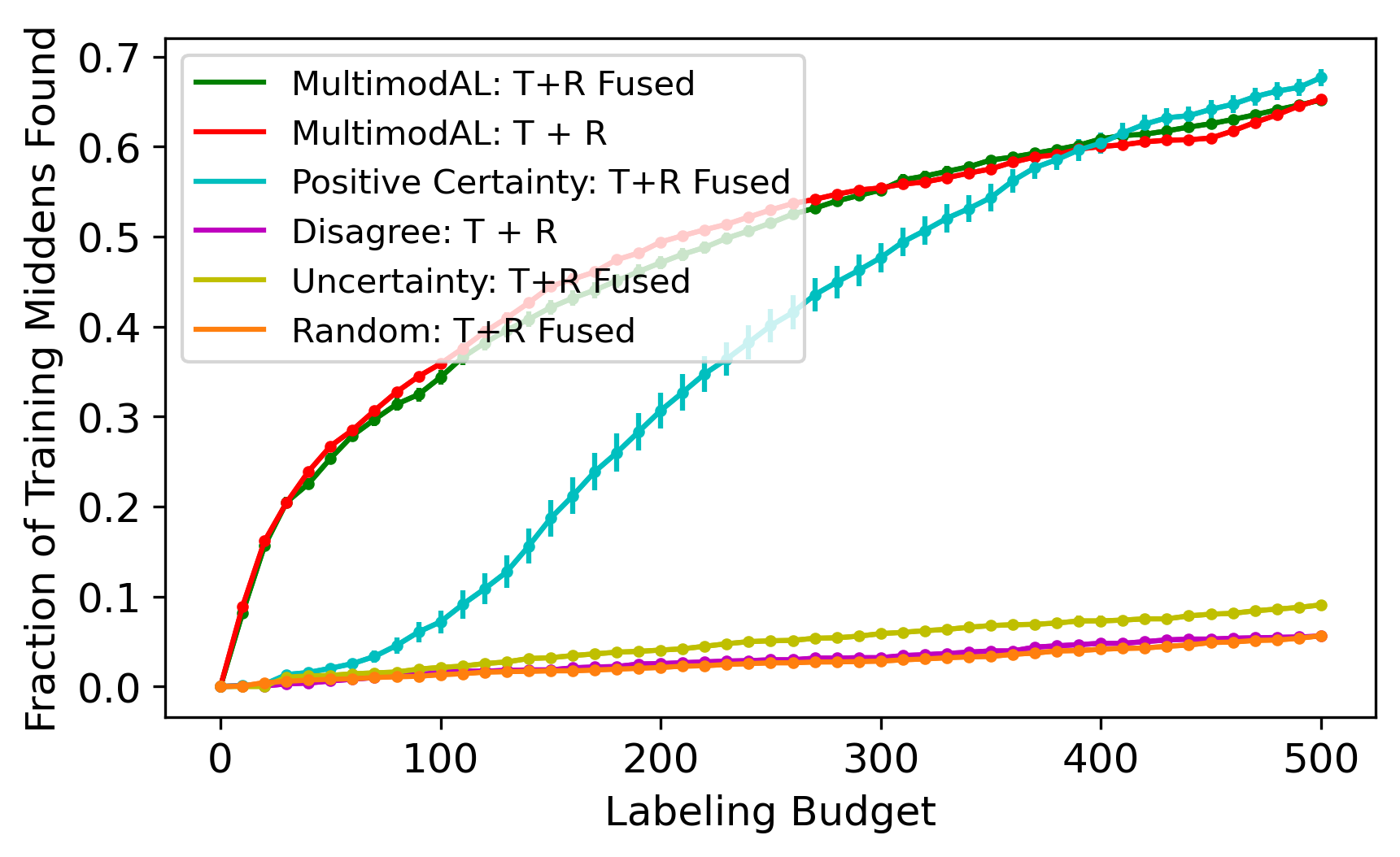}
\caption{Mean fraction of middens found in the training set for the active learning methods with up to 500 images labeled across 30 trials. Both versions of MultimodAL find more middens more quickly than the baselines with up to 400 labels.}
\label{middensfound}
\end{figure}

To dig deeper into the origin of MultimodAL's performance gains over the baselines, we consider another important evaluation metric: the fraction of middens discovered during querying by the active learning system out of all those present in the training set as more labels are provided. Figure \ref{middensfound} shows that both versions of the MultimodAL system are able to find more middens more quickly than the baselines. At around 400 labels the Positive Certainty model is accurate enough to compete with MultimodAL in terms of midden finding despite maintaining a lower test set accuracy, but for smaller numbers of labels its midden retrieval is significantly inferior. This suggests that Positive Certainty is finding less informative middens than MultimodAL, which is biased towards selecting the warmest middens. A comparison of Figures \ref{activelearning} and \ref{middensfound} indicates a positive relationship between the accuracy achieved by the active learning system and the number of positive samples it is able to find. In addition, the fact that both versions of MultimodAL find roughly the same number of middens throughout the active learning process despite MultimodAL: T + R achieving higher accuracy at the beginning is further evidence for it finding more informative middens than MultimodAL: Thermal+RGB Fused.

Figure \ref{middensfound} also demonstrates that MultimodAL is successfully addressing the class imbalance issue in our dataset. Both versions of MultimodAL find over 65\% of the middens by 500 images labeled. This is quite remarkable given that only 71 out of the 9,736 images in the training set are middens. Hence, although less than 1\% of the images given to the active learning system are middens, our active learning algorithm is able to discover 46 of them by the time 500 labels have been requested. This 9\% positivity rate is 12x larger than the overall positivity rate of middens in the training set. Perhaps more impressively, after just 100 queries MultimodAL has found over 34\% of all the middens, whereas none of the baselines manage to find more than 8\%. The number of middens found is significant because these are confirmed by humans, so we are highly confident that these are indeed middens. Once the labeling budget has been exhausted, the remaining unlabeled images are classified using the trained model, but because the model is not 100\% accurate, we have more uncertainty as to whether the images classified as middens are truly middens.

By achieving competitive accuracy and quickly identifying middens in an unlabeled dataset, MultimodAL drastically alleviates the labeling burden, as quantified in Table \ref{passive_active_comparison}. If we assume each image takes 30 seconds for a domain expert to label, then labeling every image in the dataset would require over 80 hours, but labeling 500 images would take under 4.2 hours, which is a significant amount of time saved in a resource-constrained domain --- all while identifying 65\% of all the middens in the dataset. This demonstrates the power of an active learning system that employs a domain-inspired ranking technique to find rare positive samples in a highly imbalanced dataset.

\begin{table}
\centering
\begin{tabular}{|c|c|c|c|}
\hline
    Mode & Test Accuracy & \makecell{Labeling\\Budget} & Labeling Time \\ 
\hline
    Passive & $0.86\pm0.01$ & 9,736 & 81 hours\\
\hline
    Active & $0.84\pm0.01$ & 500 & 250 mins\\
\hline
    Active & $0.81\pm0.01$ & 200 & 100 mins\\
\hline
    Active & $0.73\pm0.01$ & 50 & 25 mins\\
\hline
\end{tabular}
\caption{Comparison of passive (Thermal+RGB Fused) and active (MultimodAL: Thermal + RGB) test accuracy, labeling budget, and labeling time.}
\label{passive_active_comparison}
\end{table}

\subsection{Midden Mapping}

\begin{figure}[b]
\centering
\includegraphics[width=0.4\columnwidth]{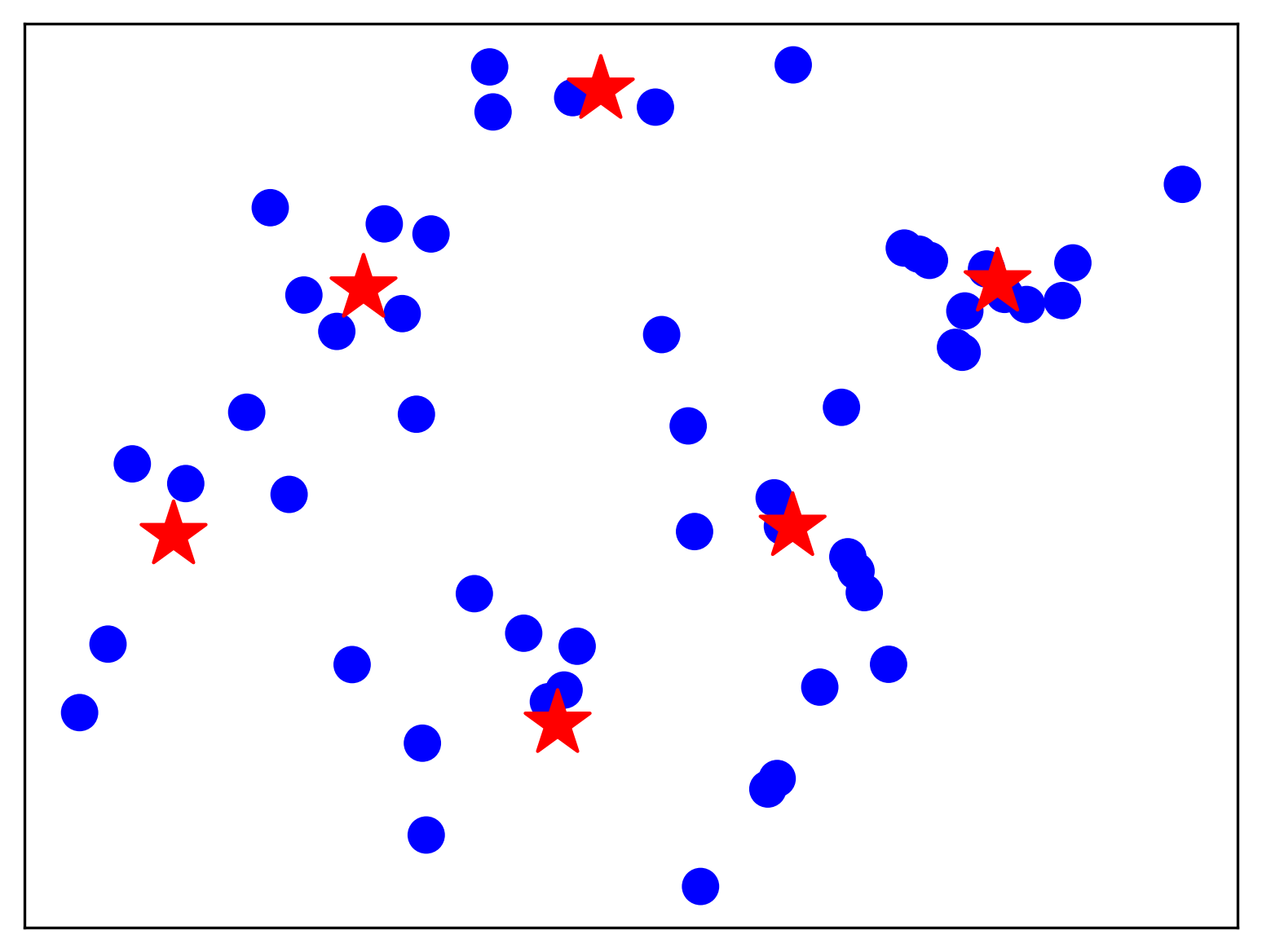}
\caption{Midden map for a 2x2-km site in Kruger National Park. Blue dots are middens. Red stars are centers of midden clusters.}
\label{middenmap}
\end{figure}

From the locations of the middens in the site, we create the first-ever midden map, presented in Figure \ref{middenmap}, and extract actionable insights. We emphasize that we are not showing landscape features on our map to protect rhino territories from poachers. Because rhino middens are used for territorial marking, by mapping these middens we can also figure out where the rhinos' territories are. We can then target ranger patrols and ecological monitors at the middens to better protect the rhinos from poachers. Unexpectedly, the midden map reveals that the middens are not distributed randomly across the region but instead form clusters. Using K-means we have identified six such clusters in our site. We recommend anti-poaching efforts prioritize those areas of high midden density to increase their efficacy.

Furthermore, we find that the rhino middens are often located along animal paths. Ecological monitors can thus focus their efforts along these paths since they are likely being used by rhinos. Additionally, knowing where the middens are is useful because monitors can visit them to more quickly find rhinos or check for fresh dung as a way of confirming the presence of rhinos in the area. This is crucial to demonstrating that the rhinos are being effectively managed and that the area should continue to be funded and resourced. Thus, by more effectively targeting scarce resources, we expect MultimodAL to have a significant impact on rhino conservation.

\section{Future Work}
In advancing a decade-long collaboration with South African National Parks, we will use MultimodAL to map middens across other sites in Kruger National Park. To do this, we will use a model pre-trained on the labeled dataset studied herein to warm-start an active learning system, further lessening the labeling burden. Unique terrain features such as termite mounds present in some other sites could increase the advantages of multimodality. To support this work, we hope to develop an interactive user interface for MultimodAL to allow domain experts to label images queried by the model(s) without needing to work directly with computer scientists. We also present two ideas for building on the MultimodAL system. First, the algorithm could be extended by dynamically weighting the maximum thermal pixel values and the model(s)' output. Second, we could define the midden detection task as one of image segmentation rather than classification and use the model from \citeauthor{segment_anything} \shortcite{segment_anything}. We also encourage the evaluation of the MultimodAL system on imagery from other domains (e.g., wildfire or animal detection).

To facilitate this future work, we make our code publicly available at \href{https://github.com/lgordon99/rhino-midden-detector}{https://github.com/lgordon99/rhino-midden-detector}. In contributing to the UN's Fifteenth Sustainable Development Goal, we encourage protected area managers to use the MultimodAL system to map rhino middens across their landscapes. However, we must keep our dataset private due to its sensitive nature with respect to the rhino poaching crisis. In accordance with the ``Leave no one behind" principle \cite{leave_no_one_behind}, we believe rhino midden maps can empower park rangers and ecological monitors in carrying out their indispensable work in conserving the planet's last wild places.

\section{Conclusion}
We make several contributions in this paper. In collaboration with South African National Parks, we map rhino middens for the first time and display where anti-poaching patrols ought to focus their efforts. Thus, we have demonstrated a way of finding and protecting rhinos without directly tracking them. We find that models passively trained on thermal, thermal+RGB fused, or thermal+LiDAR fused imagery achieve high accuracy in distinguishing between midden and empty images. To alleviate the labeling burden intrinsic to passive learning, we introduce MultimodAL, a novel active learning methodology for highly imbalanced datasets that is applicable when the instances can be ranked according to an informative metric. In addition to exploiting domain knowledge, this method also accommodates multimodal data. We demonstrate that the multi-model instantiation of the algorithm outperforms a single-model system on our dataset for a small labeling budget. Overall, both MultimodAL systems outperform baseline query strategies and find more middens than the baselines for up to 400 labels provided. Furthermore, only MultimodAL achieves an accuracy statistically indistinguishable from that of the best passive model with 94\% fewer labels, saving domain experts dozens of hours in data annotation efforts. By introducing this scalable method for mapping middens, we present new insights to support conservation practitioners in advancing rhino conservation across southern Africa.

\section*{Acknowledgments}
Research was sponsored by the DARPA and was accomplished under Grant Number: HR001122C0182. The views and conclusions contained in this document are those of the authors and should not be interpreted as representing the official policies, either expressed or implied, of DARPA or the U.S. Government. The U.S. Government is authorized to reproduce and distribute reprints for Government purposes notwithstanding any copyright notation herein. J.A.K. was supported by an NSF Graduate Research Fellowship under grant DGE1745303. South African National Parks is thanked for logistical and scientific support, as well as permission to work in Kruger National Park. All drone flights were performed with approval from the South African Civil Aviation Authority and South African National Parks Airwing.

\bibliographystyle{named}
\bibliography{references}

\end{document}